\documentclass[runningheads]{llncs}
\usepackage{graphicx}
\usepackage[linesnumbered,ruled,vlined]{algorithm2e}
\usepackage{algpseudocode}
\usepackage{float}
\restylefloat{table}
\usepackage{tabu}
\begin{document}
\title{Determining Optimal Number of k-Clusters based on Predefined Level-of-Similarity}
\titlerunning{Clustering algorithm based on Predefined Level-of-Similarity}
%
\author{Rabindra Lamsal \and
Shubham Katiyar}
\authorrunning{Lamsal \& Katiyar}
%
\institute{School of Computer and Systems Sciences\\ Jawaharlal Nehru University, New Delhi 110067\\
\email{rabindralamsal@outlook.com, shubha25\_scs@jnu.ac.in}}
\maketitle
\begin{abstract}
This paper proposes a centroid-based clustering algorithm which is capable of clustering data-points with n-features, without having to specify the number of clusters to be formed. The core logic behind the algorithm is a similarity measure, which collectively decides whether to assign an incoming data-point to a pre-existing cluster, or create a new cluster and assign the data-point to it. The proposed clustering algorithm is application-specific and is applicable when the need is to perform clustering analysis of a stream of data-points, where the similarity measure between an incoming data-point and the cluster to which the data-point is to be associated with, is greater than the predefined Level-of-Similarity.

\end{abstract}

\section{Introduction}
The main gist behind clustering is to group data-points into various groups (clusters) based on their features, i.e. properties. The generation of clusters varies application-wise \cite{survey}, because it depends on what factors are to be taken into consideration to form a particular cluster. But, the focus of every clustering algorithm remains same, i.e. to group similar data-points to a common cluster. The thing that differs is how this goal of forming a cluster is achieved. Different algorithms use different concepts to deal with similarity measure among the data-points. There are many popular clustering algorithms that group data-points based on various strategies to define the similarity measure between them. Centroid-based \cite{centroid}, density-based \cite{den}, graph-based \cite{graph}, etc. are the commonly used strategies.

Often used algorithms like k-means, hierarchical clustering, DBSCAN, etc. require a set of data-points in space beforehand. Without making fine adjustments to these pre-existing clustering algorithms, it is not possible to cluster a stream of near real-time data-points. Therefore, a real-world problem exists if there is a necessity of grouping a stream of data-points without having to specify the number of clusters to be generated.

Therefore, taking this limitation into consideration, this paper proposes a machine learning algorithm that can group incoming data-points without having to specify the number of clusters to be formed. The main agenda behind proposing this algorithm is to facilitate those problems which require clustering of data-points based on some level-of-similarity.

The paper is organized as follows. Section 2 describes the core logic of the proposed clustering algorithm. Implementation of algorithm is done in Section 3 and Section 4 is the conclusion.

\section{The Proposed Clustering Algorithm}
This algorithm requires an initial declaration of cluster strictness. Cluster Strictness is the lowest permitted level-of-similarity between a data-point and the centroid of a cluster. Here, cluster centroid is the average of features of the data-points present in a cluster and is an effective way of representing a particular cluster. If cluster strictness is set to a measure of 60, then the accepted variability between a cluster centroid (C) and a data-point (D) should be a measure of 40 at-most, for D to be associated to cluster C.

Hence, this algorithm is capable of generating clusters based on a level-of-similarity. If one needs clusters with very low variance among the data-points, the cluster strictness can be adjusted accordingly; maybe around 80-95\%. That is why cluster strictness plays a significant role in generating clusters, and its value depends entirely on what application the clustering is being done.

\subsection{Data-point and Features}
A data-point can have multiple features (n-features). Those features are the characteristics which collectively define a data-point. A stream of data-points can be grouped into various clusters, based on their features. Below is an example of a data-point with 7 features.
\begin{center}
A data-point:	5	10	15	20	25	30	35
\end{center}
During the implementation of the proposed algorithm, the above pattern is used to represent a data-point, and similarity measure between the data-points and cluster(s) is calculated accordingly. The algorithm takes data-points with n-features, one data-point at a time, and requires all the data-points to have the same fixed number of features. When the algorithm runs for the very first time and is waiting for a new data-point; the number of cluster(s) is 0. Cluster strictness needs to be defined beforehand. Based on the requirements, this value can be adjusted. Higher the value of cluster strictness, less is the variance between the data-points in a cluster. Depending on the data-points, using a higher value of cluster strictness result in a large number of clusters. This is more likely to happen if the incoming data-points are less identical to each other.

After assigning some value to cluster strictness, say 70\%, the number of features that should be matched between a data-point and a cluster is calculated using the formula given below:

\begin{equation}
should\_match\_features = \frac{no\_of\_features * cluster\_strictness}{100}\;
\end{equation}

Suppose, if the data-points that are to be clustered have 20 features. That means cluster strictness with 70\% results into 14 features that must be matched (at least) for a data-point to be associated with a particular cluster.

\subsection{Qualifying Feature}
A feature qualifies to be counted as a matched feature when its similarity measure lies between the range: (cluster\_strictness) and (100 + (100 – cluster\_strictness)). That is, the valid range is 70-130 when cluster strictness is considered 70. Suppose, when variability check of 5 and 13 is to be done with respect to 9, basic mathematics can be used.  9 with respect to 9 gives a similarity measure of 100. 5 with respect to 9 gives a similarity measure of 55.56. And, 13 with respect to 9 gives a similarity measure of 144.44. If allowed variability is considered to be a measure of 50, then similarity measure of both 5 and 12 falls under the range 50-150, and are considered to be associated to 9 by at-least a measure of 50.

In the case of data-points, the following formula is used to calculate the similarity measure.

\begin{equation} S(Ci,Fj)) = \frac{100 * datapoint_j}{centroid(i,j))}
\end{equation}

\begin{center}
\textit{where i is an index for clusters, and j is an index for features.}
\end{center}

As an example, in formula 2, i=3 and j=7 reflect the fact that cluster 3 is being considered, and 7th feature of a data-point and 7th feature of the centroid of cluster 3 are being used to compute the level-of-similarity between them. If the resulted similarity measure lies between the range (cluster\_strictness) and (100 + (100 – cluster\_strictness)), then the 7th feature is said to be matched, and the value of matched feature counter for the cluster 3 is incremented by 1. This process of calculating the similarity feature of an incoming data-point is done with all the cluster(s). And if the matched feature counter for a cluster reaches the limit of the minimum number of features that should be matched between a data-point and a cluster, that particular cluster is now placed into a list of qualified clusters.

\begin{algorithm}
\caption{Clustering Algorithm for Data-points with n Features}
\SetAlgoLined
\DontPrintSemicolon

Declare cluster\_strictness.\;
Initialize $cluster\_counter \gets 0$.\;
Calculate should\_match\_features using formula.\;
Read a datapoint \(D\) \;
\eIf{cluster\_counter = 0}{
Increment cluster\_counter by 1.\;
Create a new cluster C\textsubscript{cluster\_counter}.\;
Assign \(D\) to the newly formed cluster.\;
\(D\) be the centroid of the cluster.\;
}
{

\For{all clusters C\textsubscript{i}, where i ranges from $1$ to cluster\_counter}{
    \For{all features F\textsubscript{j}, where j ranges from $1$ to no\_of\_features}{
    Calculate similarity\ measure using formula. \;
\If{S(Ci,Fj) \textgreater= (cluster\_strictness)\; \quad \quad and S(Ci,Fj) \textless=(100 + (100 - cluster\_strictness))} {
Increment matched\_features\textsubscript{i} by 1.\;
}
  }
\If{matched\_features\textsubscript{i} \textgreater= should\_match\_features} {
Add C\textsubscript{i} to the list of qualified\_clusters\;
} 
  }

\If{qualified\_cluster list is empty}{ 
Increment cluster\_counter by 1.\;
Create a new cluster C\textsubscript{cluster\_counter}.\;
Assign \(D\) to the newly formed cluster.\;
\(D\) be the centroid of the cluster.\;}

\eIf{qualified\_cluster list contains single cluster}{ 
Assign \(D\) to the cluster.\;
Re-calculate the centroid.\;}
{

Calculate cluster(s) with maximum matched\_features.\;
\eIf{single cluster arises}{
Assign \(D\) to the cluster.\;
Re-calculate the centroid.\;}
{
Find the cluster with max average of qualifying similarity measures.\;
Assign \(D\) to the cluster.\;
Re-calculate the centroid.\;
}
}
}
\end{algorithm}

\subsection{Qualified Cluster List}
This list contains the cluster(s) that satisfy the condition of having at least a minimum number of features that have matched with an incoming data-point. After calculation of similarity measure for a data-point with all the cluster(s), there might arise any one of the following three situations.

\subsubsection{A qualified cluster list is empty}
The qualified list being empty indicates that no similar cluster(s) were found within the provided level-of-similarity. Hence, a new cluster is generated, and the data-point is assigned to the newly formed cluster. The data-point is the centroid of this cluster.

\subsubsection{A Qualified cluster list contains exactly 1 cluster}
In this case, the data-point is simply assigned to the cluster which is present in the list. And, the centroid of the cluster is re-calculated.

\subsubsection{A Qualified cluster list contains more than 1 cluster}
Case 1: If the list contains more than 1 cluster, the cluster with maximum matched features is identified. The data-point is now assigned to the identified cluster. Case 2: Sometimes, two or more cluster might come-up with the same highest number of matched features. In this case, the cluster with a maximum average of qualifying similarity measures is identified, and the data-point is assigned to that cluster accordingly.

\subsection{Conflicting Clusters}
Whenever multiple clusters show up in qualified cluster list, while trying to identify the nearest similar cluster based on the highest number of matched features, those clusters are said to be the conflicting ones. In such case, an average of qualifying similarity measures is calculated for all the clusters which are in the qualified list and have the same number of matched features, and finally, the cluster with the maximum average is considered to be the cluster to which the data-point is associated. While calculating an average, only the qualifying similarity measures are being considered. One fact is also being taken into consideration that technically 60 and 140 are same with respect to 100. Both have a variability measure of 40.

\section{Implementation of the Proposed Algorithm}
Let us consider the following six data-points, each of 10 features, that are to be clustered. The data-points are input to the algorithm in a serial fashion.\\\\
\begin{tabu} to 1\textwidth { |c|X[c]|X[c]|X[c]|X[c]|X[c]|X[c]|X[c]|X[c]|X[c]|X[c]| }\hline
Data-point 1& 10&	15&	20&	25&	30&	35&	40&	45&	50&	55\\\hline
Data-point 2& 09&	35&	18&	45&	10&	32&	60&	41&	10&	20\\\hline
Data-point 3& 18&	13&	18&	27&	30&	38&	38&	41&	49&	57\\\hline
Data-point 4& 20&	20&	18&	05&	15&	34&	50&	43&	10&	50\\\hline
Data-point 5& 17&	17&	18&	15&	22&	35&	44&	43&	10&	53\\\hline
Data-point 6& 10&	32&	20&	45&	12&	55&	40&	55&	09&	25\\
\hline
\end{tabu}\\\\

Let us suppose, level-of-similarity for a cluster is considered to be 60. That means variability measure of 40 between a data-point and a cluster centroid is acceptable. With cluster strictness set to 60, using formula 1 we get minimum number features that must be matched to be 6.

Initially, when Data-point 1 is input to the algorithm, because of the absence of cluster(s), a new cluster C1 is created, and Data-point 1 is assigned to it, and features of Data-point 1 is assumed to be the centroid of C1.
\begin{center}
C1 = Data-point 1\\
C1: 10	15	20	25	30	35	40	45	50	55
\end{center}

For Data-point 2\\\\
\begin{tabu} to 1\textwidth { |c|X[c]|X[c]|X[c]|X[c]|X[c]|X[c]|X[c]|X[c]|X[c]|X[c]| }\hline
S(C1,Fi)& 90&	233.34&	90&	180&	33.34&	91.43&	150&	91.11&	20&	36.36\\
\hline
\end{tabu}\\

The number of qualified features is 4, which is less than 6. C1 cannot be added to the list of qualified clusters. Since there are no clusters in the list of qualified clusters, a new cluster C2 is generated, and Data-point 2 is assigned to C2.

\begin{center}
C2 = Data-point 2\\
C2: 09	35	18	45	10	32	60	41	10	20
\end{center}

For Data-point 3\\\\
\begin{tabu} to 1\textwidth { |c|X[c]|X[c]|X[c]|X[c]|X[c]|X[c]|X[c]|X[c]|X[c]|X[c]| }
\hline
S(C1,Fi) & 180&	86.67&	90&	108&	100&	108.57&	95&	91.11&	98&	103.64\\\hline
S(C2,Fi) & 200&	37.14&	100&	60&	300&	118.75&	63.33&	100&	490&	285\\
\hline
\end{tabu}\\

Only C1 qualified to be in in the list of qualified clusters, Data-point 3 is assigned to C1.

\begin{center}
C1 = Data-point 1, Data-point 3\\
New C1:	14	14	19	26	30	36.5	39	43	49.5	56
\end{center}

For Data-point 4\\\\
\begin{tabu} to 1\textwidth { |c|X[c]|X[c]|X[c]|X[c]|X[c]|X[c]|X[c]|X[c]|X[c]|X[c]| }\hline
S(C1,Fi) & 142.86&	142.86&	94.74&	19.23&	50&	93.15&	128.21&	100&	20.20&	89.29\\\hline
S(C2,Fi) & 222.22&	57.14&	100&	11.11&	150&	106.25&	83.33&	104.87&	100&	250\\
\hline
\end{tabu}\\

Since there no cluster qualified to be kept in the list of qualified clusters, a new cluster C3 is generated, and Data-point 4 is assigned to C3.

\begin{center}
C3 = Data-point 4\\
New C3:	20	20	18	05	15	34	50	43	10	50
\end{center}

For Data-point 5\\\\
\begin{tabu} to 1\textwidth { |c|X[c]|X[c]|X[c]|X[c]|X[c]|X[c]|X[c]|X[c]|X[c]|X[c]| }\hline
S(C1,Fi) & 121.43&	121.43&	94.74&	57.59&	73.33&	95.89&	112.82&	100&	20.20&	94.64\\\hline
S(C2,Fi) & 188.89&	48.57&	100&	33.33&	220&	109.38&	73.33&	104.88&	100&	265\\\hline
S(C3,Fi) & 85&	85&	100&	300&	146.67&	102.94&	88&	100&	100&	106\\
\hline
\end{tabu}\\

Since C1 and C3 qualified to be in the list of qualified clusters and both the clusters have the same number of qualified features. Now, the average of the qualifying features is calculated for both the clusters. Qualifying features beyond 100 are scaled below 100 using the formula given below:

\begin{equation}
scaled\ value = 100 - (value\ beyond\ hundred - 100)\;
\end{equation}

\begin{center}
Average of qualifying features (C1) = 87.87\\
Average of qualifying features (C3) = 93.63
\end{center}

Since the average of qualifying features for C3 is higher, Data-point 5 is assigned to C3.

\begin{center}
C3 = Data-point 4, Data-point 5\\
New C3:	18.5	18.5	18	10	18.5	34.5	47	43	10	51.5
\end{center}

For Data-point 6\\\\
\begin{tabu} to 1\textwidth { |c|X[c]|X[c]|X[c]|X[c]|X[c]|X[c]|X[c]|X[c]|X[c]|X[c]| }\hline
S(C1,Fi) & 71.43&	228.57&	105.26&	173.08&	40&	150.68&	102.56&	127.91&	18.18&	44.64\\\hline
S(C2,Fi) & 111.11&	91.43&	111.11&	100&	120&	171.88&	66.67&	134.15&	90&	125\\\hline
S(C3,Fi) & 54.05&	172.97&	111.11&	450&	64.86&	159.42&	85.11&	127.91&	90&	48.54\\
\hline
\end{tabu}\\

Since C2 is the only cluster qualifying to be in the list of qualified clusters, Data-point 6 is assigned to C2.

\begin{center}
C2 = Data-point 2, Data-point 6\\
New C2:	9.5	33.5	19	45	11	43.5	50	48	9.5	22.5
\end{center}

The data-points were successfully grouped into 3 various clusters based on the similarity measure of their features. Cluster strictness of 60 was considered, hence permitted variability measure between a data-point and a cluster centroid was 40.

\section{Conclusion}
The theoretical aspect of the proposed algorithm was discussed and implemented on 6 data-points (all of them with 10-features) with the supposition that they are input to the algorithm one after another, just like a stream-of-data in a serial fashion. The formation of clusters could be manipulated by increasing/decreasing the level-of-similarity. The algorithm is applicable in a real-world scenario where the task is to group a stream of data-points, incoming to a system, based on their similarity with the average of features (centroid) of data-points existing in their respective previously formed clusters. If none of the existing clusters satisfy the similarity measure for a new data-point, a new cluster is formed, and the data-point is assigned to the newly formed cluster.

When the number of clusters in the space is relatively large, the complexity of the algorithm increases proportionately. Therefore, it would be an interesting problem to try decreasing the number of similarity checks that is done when a new data-points arrives. Instead of checking the similarity of data-points with all the pre-existing clusters, an approach can be made so that only a certain number of clusters are taken into consideration for similarity check. This way, the number of required computations can be decreased.

\section*{Acknowledgment}
The authors would like to gratefully acknowledge Intel for providing computing cluster during this study.

\end{document}